# Computer Vision based Automated Quantification of Agricultural Sprayers Boom Displacement

Aryan Singh Dalal[*1], Sidharth Rai[2], Rahul Singh[3], Treman Singh Kaloya[4], Rahul Harsha Cheppally[5], and Ajay Sharda[*6]

[*1]*Dept. of Computer Science, Kansas State University, 66506, Manhattan, KS, USA*

[2]*College of Business, Kansas State University, 66506, Manhattan, KS, USA*

[3]*Dept. of Biological and Agriculture Engineering, Kansas State University, 66506, Manhattan, KS, USA* [4]*Dept. of Biological and Agriculture Engineering, Kansas State University, 66506, Manhattan, KS, USA* [5]*Dept. of Biological and Agriculture Engineering, Kansas State University, 66506, Manhattan, KS, USA*

[*6]*Dept. of Biological and Agriculture Engineering, Kansas State University, 66506, Manhattan, KS, USA*

aryand@ksu.edu

asharda@ksu.edu

**Abstract**: Application rate errors when using self-propelled agricultural sprayers for agricultural production remain a concern. Among other factors, spray boom instability is one of the major contributors to application errors. Spray booms' width of 38 m, combined with 30 kph driving speeds, varying terrain, and machine dynamics when maneuvering complex field boundaries, make controls of these booms very complex. However, there is no quantitative knowledge on the extent of boom movement to systematically develop a solution that might include boom designs and responsive boom control systems. Therefore, this study was conducted to develop an automated computer vision system to quantify the boom movement of various agricultural sprayers. A computer vision system was developed to track a target on the edge of the sprayer boom in real time. YOLO V7, V8, and V11 neural network models were trained to track the boom's movements in field operations to quantify effective displacement in the vertical and transverse directions. An inclinometer sensor was mounted on the boom to capture boom angles and validate the neural network model output. The results showed that the model could detect the target with more than 90 percent accuracy, and distance estimates of the target on the boom were within 0.026 m of the inclinometer sensor data. This system can quantify the boom movement on the current sprayer and potentially on any other sprayer with minor modifications. The data can be used to make design improvements to make sprayer booms more stable and achieve greater application accuracy.

# INTRODUCTION

Agricultural land is limited, and with the increase in population, it is becoming increasingly difficult to expand farming operations [1]. Innovative practices that utilize resources efficiently and increase crop yields are needed to feed this ever-growing population. One of the methods to increase crop yield is to decrease the detrimental effect of pests and weeds on the crop [2]. Spraying is essential as it is pivotal in regulating pests while enhancing crop yield and quality. Over the last decade, self-propelled sprayer technologies have evolved by integrating pulse width modulation (PWM) rate controllers, environmental sensors, auto rinse systems, and boom height control. Most technology enhancements have been implemented to manage target application rates during dynamic and diverse application scenarios like irregularly shaped fields, curvilinear sprayer operation, speed variations, and spray control-section actions. Application rates are significant because overapplication increases the danger of environmental damage and excessive residues, while underapplication can result in partial pest control. Correct chemical application rates result in effective pest control and minimize the economic impact of uneven application [3]. However, traditional self-propelled sprayers tend to provide inaccurate chemical applications on the crops. It has been noted that most of the time, only 2% of such chemicals reach their intended target, while the rest pollute soils, surface water, and groundwater [1, 3, 4].

Though flow rate controllers have significantly reduced application errors (Jonathan Cite), one of the main contributors to inaccurate spraying is the uncontrolled movement of the spray boom. During sprayer operation on uneven terrain, a small sprayer radius of turning around field corners and grassed waterways, and rapid acceleration and deceleration on headlands, these spray booms encounter various vibrations, including roll and yaw, which typically lead to unintended movements. During yaw and roll changes, spray booms can move significantly, causing undesirable spraying operations [5]. The vertical movements of a boom can substantially impact the uniformity of spray droplet deposit density both longitudinally and laterally along the vehicle's track. This is due to the change in nozzle-to-nozzle overlap and spray droplet coverage (droplet density) due to the change in the nozzle height on the boom from plants. Furthermore, horizontal fluctuations in the velocity of the boom can also cause spray droplet coverage variations along the track. The primary source of boom movements is attributed to the frame movements of the sprayer. However, a suspension system is typically implemented between the boom and the frame to mitigate these undesired movements. The suspension is designed to isolate the boom from the roll motion of the sprayer frame and thus minimize the occurrence of unwanted boom movements. This is the most crucial component in reducing the potential impact of boom movements on the uniformity of coverage during application [5] and reducing potential detrimental effects [5]. Even after the implementation of the suspension system, considerable concerns around the amount of movement still exist, which shows that the suspension system is not fully capable of eliminating the boom's movements.

Research efforts spanning over two decades have been dedicated to advancing the development and implementation of precision spraying techniques, thus contributing to the state-of-the-art methods used in this practice today. Many studies have been done to quantify boom movement and thereby measure its outcomes. The evaluation techniques employed to assess boom movements and deformations in field conditions can be categorized into two groups: absolute and relative measurement methods. Among these methods, the most commonly used absolute approach involves the use of small seismic accelerometers attached to the sprayer. This technique provides the advantage of not altering the boom's behavior during measurement. Additionally, it can be easily implemented on any sprayers currently in operation in the field, allowing for comparisons between them [6]. Some researchers [7] developed a method to assess height variations of boom sprayers by computing displacements through double integration of accelerometer data. However, this approach faces challenges in determining the cutoff frequency of the appropriate filter required to eliminate any long-term trends or inherent drifts in accelerometer outputs. An alternative approach to address the limitations of the accelerometer method is the semi-automated method, which utilizes an ultrasonic sensor [8]. However, due to the sensor's limited scanning range, this method only provides data over short distances (5-10 m). Additionally, other researchers proposed a method based on a fixed laser sensor [9], but this approach is also limited to measuring over a restricted distance. Relative displacement methods, on the other hand, use the vehicle as the reference point. Some other authors proposed a technique that utilized a laser distance meter mounted on the boom. It directed its beam towards a specially designed reflective target fixed rigidly at the front of the tractor and oriented perpendicular to the beam [10]. However, the main disadvantage of this method is its tediousness due to the requirement for precise positioning of the target before conducting trials [10].

Experiments have also been done in a laboratory utilizing a machine composed of a movable carriage and a section of a sprayer boom. The sprayer boom could perform a sinusoidal motion for every meter of travel at an average velocity of 1 m/s, while water and fluorescent dye were used for spraying. The fluorescent dye concentration was measured using spectrophotometry, with specialized filter papers measuring 10 x 10 cm placed on a collecting surface [6]. However, the primary constraint of the method was the inability to replicate the boom's movements in the field precisely. Due to the imposed simple sinusoidal movements on the boom component, extrapolating the results to estimate the actual spray distribution from an existing machine posed significant difficulties.

Researchers conducted a study to investigate the impact of horizontal movements of a sprayer boom on the longitudinal spray distribution in field conditions. The study combined field and laboratory aspects and focused on analyzing the effects of the boom horizontal movements on deposition onto artificial targets. The boom movements of a sprayer were measured in the field using a specially designed measurement chain, and the resulting ground distribution was estimated by spraying a dye (Nigrosine) on large sheets of wallpaper and using image analysis to compute the spray coverage. The study results showed that the sprayer boom's horizontal movements largely influence the spray deposit longitudinal distribution, even if other parameters such as boom height variations and wind effect are present. However, the spray distribution variation was lower than the horizontal speed variation due to the filtering effect of the spray pattern. The amplitude of the boom movements was investigated across an extensive range of situations, including five different sprayers and three different crops. Based on the measured movements and the computed RCV, the variability of the longitudinal spray distribution due to horizontal movements was estimated at between 3.5% and 5.2% [11]. Nevertheless, it is essential to note that this method's scope may be restricted since its effect assessment only covers horizontal inclination. Overall, no robust system can measure horizontal and vertical boom movement in real time during actual field operating conditions. Such a system is especially needed to understand the magnitude and frequency of boom movement in current commercial sprayers, so corrective design measures should be implemented to address those concerns.

Computer science and Machine Learning (ML) are increasingly used in agricultural applications to understand the performance of current machine systems and quantify operational gaps. Machine learning is a core field of artificial intelligence, enabling computers to learn independently by utilizing data to forecast or make decisions without explicit programming. Examples of ML use include agricultural yield prediction, disease detection and monitoring, irrigation management, soil analysis, planting systems, sprayers, and autonomy [12, 13, 14, 15, 16, 17, 18]. Additionally, agricultural pests and diseases can be detected using machine learning, enabling farmers to take action to prevent or mitigate the harm. Machine learning algorithms are being developed to identify and classify various weeds and crops, enabling focused spraying operations that reduce pesticide loss to the environment, enable site-specific applications, optimize pesticide use, and enhance profitability. Therefore, this study was designed to develop a computer-vision-based solution that can quantify the vertical and fore-and-aft movement of booms on self-propelled sprayers. The designed study had two specific objectives: 1) develop a computer vision system to quantify boom movements during actual field conditions and 2) validate the system performance through both lab simulations and during self-propelled sprayer operation in an agricultural field.

## METHODOLOGY

### Parameter Considerations

#### *Aruco Marker*

The approach for determining the displacement of the boom using a computer-vision system was twofold. Firstly, detect a specifically designed object mounted at the tip of the boom, and secondly, track the selected object in real-time while the boom is in motion. To accomplish object detection, Aruco Markers were employed (Figure 1. Aruco Marker with 6*6 by dictionary and ID no 3), which were synthetic square markers that were part of the markers and dictionary of the OpenCV library. The synthetic square markers were characterized by a large black border surrounding an inner binary matrix, which functions as a unique identifier (id) for the marker. The binary code embedded within the marker enables its identification and supports the use of error detection and repair algorithms.

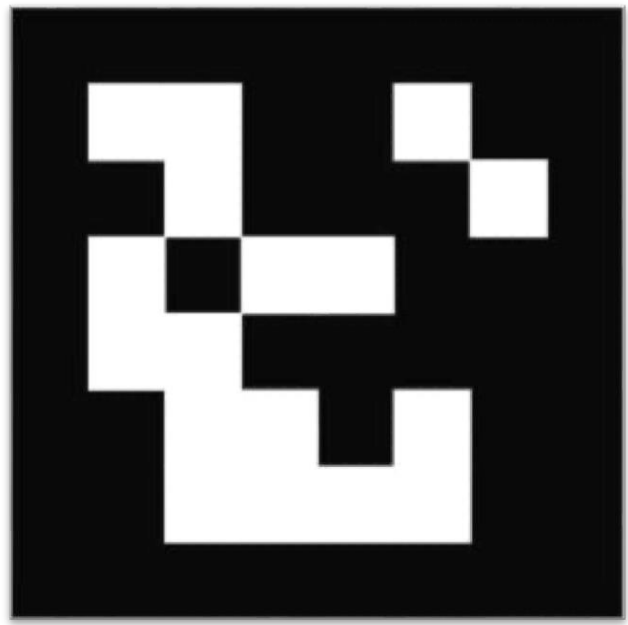

*Figure 1. Aruco Marker with 6*6 by dictionary and ID no 3*

The marker's black border enhances its visibility within an image. The size of the inner matrix was determined by the marker size, whereby a marker size of 4x4 comprises 16 bits.

Aruco marker detection is a two-phase process designed to identify specific image patterns. The first phase involves analyzing the image to locate potential markers, which are typically square-shaped. This is achieved through image-processing techniques that highlight distinct shapes and filter out irrelevant elements based on size, form, and proximity. Once potential markers are identified, the second phase begins, which involves a detailed examination of each candidate marker's internal structure. The image of each potential marker is adjusted to provide a clear, front-on view. Then, it's converted into a distinct black and white pattern and divided into a grid. The resulting arrangement of black and white cells is compared against a pre-established set of valid patterns. If necessary, error correction methods are applied to account for slight discrepancies. This comprehensive approach ensures accurate identification of genuine Aruco markers within the image, balancing efficiency with precision.

An ArUco marker, measuring 0.42 ×0.29 m, was mounted on the underside of the left boom of a New Holland Guardian SP410F self-propelled sprayer. The sprayer, equipped with a 36.5-meter-wide boom, was employed for the study in tracking and data collection (Figure 2. Field environment for Aruco Marker). In the context of detecting the Aruco marker at the end tip of the boom in a video captured from a distance of 18.2 meters, several factors were found to hinder its detection. One of the most prominent factors was the extended distance between the camera and the object of interest. Due to this distance, the size of the Aruco marker appeared small, resulting in difficulties in detecting it accurately.

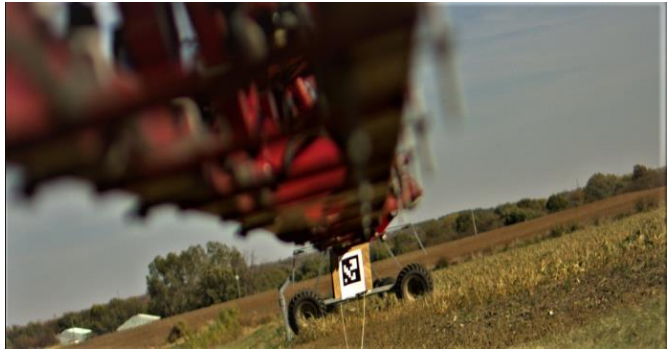

*Figure 2. Field environment for Aruco Marker*

Additionally, the size of the Aruco marker (Size: 0.58 by 0.42 meters) could not be increased beyond a certain limit due to practical limitations, such as the weight and balance of the boom. Furthermore, the use of a larger marker could potentially affect the movement of the boom, thereby leading to errors in detection.

Another factor that affected the detection was the interference of the crop during the growing season. The Aruco marker could be obstructed by the crops, resulting in the marker being hidden from view, thereby impeding its detection. Moreover, attempting to zoom in on the frame did not resolve this issue, as the Aruco marker moved out of the frame while the sprayer was in motion. These observations exhibited that for the application of tracking objects placed at a larger distance from the imagery system, the Aruco marker could not be used for the purpose of object detection.

***Machine Learning***

Following a comprehensive evaluation of the restrictions and inadequacies inherent in the prior methodology, an alternate strategy was adopted. The present methodology involved the use of a neural network, a variant of artificial intelligence modeled after the neurological structure of the human brain. The present age observes a substantial utilization of neural networks in the realm of object detection. At the time this research project began, YOLOv7 was the most recent neural network algorithm used for object identification. The YOLO version 7 neural network option had garnered significant attention for its outstanding performance in the Mean Average Precision (MAP) score metrics related to object detection, thereby positioning it as the primary preference among other available neural network alternatives [19, 20, 21]. The impressive achievement of this neural network can be attributed to its proficient and effective utilization of deep convolutional neural networks (CNNs), facilitating a heightened degree of precision in object detection in real-time scenarios. One of the primary factors contributing to the superior precision of YoloV7 in the domain of object detection, compared to other CNNs, is attributed to its distinctive architectural design. The process of feature extraction in the YoloV7 model commences with the initial passage of image frames through a backbone. This phase involves retrieving pertinent features from the images. Subsequently, these distinctive characteristics are amalgamated and merged within the neck of the network. Subsequently, the amalgamated attributes are conveyed to the apex of the network, where YOLO predicts the localizations and categorizations of entities in the visual representation, and generates bounding boxes encircling them. In addition to its primary methodology, YOLO utilizes a post-processing approach known as non-maximum suppression (NMS) to develop its ultimate predictions. The present technique is employed to eliminate numerous concurrently generated bounding boxes for a single object and retain only the bounding box characterized by the highest confidence score [21] . This approach enhances the precision of object detection by reducing the occurrence of unnecessary bounding boxes. To align the research work with the given methodology, the first and foremost step was to select the target.

### *Target Selection*

Advancing in object detection involved finalizing the target object, serving as a proxy for detection. Various factors were considered in determining the target, including its shape and color. Additionally, it was crucial to maintain a small size for the target to mitigate similar limitations experienced with the Aruco marker. The shape of the target drew inspiration from the distinct form of a multi-cellular organism's cell known as the "Scutoid" (Figure 3. Target Object). Emphasizing a distinct shape for the target was deemed crucial to prevent any inadvertent resemblance to existing elements in the field. This measure aimed to avoid potential confusion for the detection algorithm, ensuring accurate and reliable identification of the designated target. Similar criteria were applied when finalizing the color elements of the shape; the color elements of the target shape should differ from those present in the field. This necessitated the use of multiple colors in the finalized shape. It was also considered that the chosen color combination should remain visible in various lighting conditions and backgrounds in the field.

### *Machine Vision System*

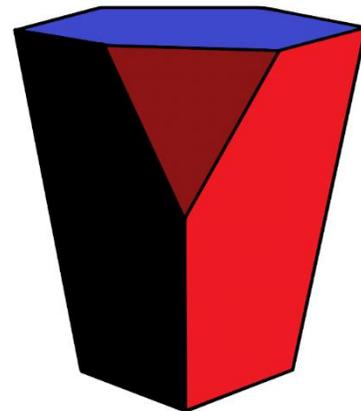

The camera selection process was conducted with careful consideration of the dynamic movement of the boom while the sprayer is in motion. Critical factors such as focal length and clarity were identified as key criteria in the camera selection process, with a particular emphasis on achieving optimal image quality at long distances. The selection process ultimately led to the identification of the Basler acA1920-50gc color camera, which offers high-resolution data acquisition capabilities at 1920*1200 pixels and 2.3 megapixels, with a maximum frame rate of 50 frames per second. A 35mm ruggedized Kowa lens (LM35HC-V | 1" Ruggedized 35mm 5MP C-Mount Lens) was paired with the Basler camera. Multiple factors were taken into account, such as the desired focal length and resolution needed for capturing clear and detailed images of the boom's motion while in operation. The ruggedized design of the lens was also an important consideration, ensuring durability and reliability during use in potentially challenging and demanding agricultural environments. Overall, the camera-lens combination achieved high-quality data. An appropriate mounting location of the camera system was required to effectively capture the desired target. The camera was enclosed in an aluminum enclosure (CH50.DIA50.PMMA, Argutec, Ostrava, Czech Republic) to provide an IP65 rating while operating in the field and to maintain an uninterrupted view of the tip of the boom.

*Figure 3. Target Object selected for detection*

### *Experimental Boom – Lab Calibration and Validation*

A specialized boom structure was developed for the laboratory context to recreate a comprehensive array of boom movements, particularly in challenging scenarios (Figure 4. Experimental Boom). This boom stand was constructed to replicate diverse boom movements in pre-defined horizontal and vertical directions. The image capturing system, along with the neural network model to detect selected target, was used to develop a calibration process for boom displacement. Several design considerations were implemented to ensure a thorough and precise experimentation process. A 3-meter-long boom section was seamlessly integrated into the central segment of the boom stand, extending backward to act as a supportive counterweight. This boom section featured articulate joints, enabling controlled movements along the horizontal axes for precise manipulation during tests. Simultaneously, an actuator (LA31 CARELINE, LINAK, Louisville, USA) was strategically attached at the boom's initiation point to facilitate vertical elevation, granting the capacity for motion along the vertical axes. The boom section can be moved in 3 m horizonal plane and more than 1.2 meter in vertical plane in total. A laboratory boom was designed to aid in the rapid development of the detection model and the validation of the system. This boom consists of a base stand of height 1.7 meters and an extended arm (boom), of length 3 meters. The developed boom could move 10 meters on the horizontal axis and more than 1.2 meters on the vertical axis in both up and down positions. This boom was used to simulate controlled movement in both horizontal and vertical motion to determine the error in the inclinometer sensor (TMS/TMM22, SICK AG, Waldkirch, Germany), attached at the beginning of the boom. Such movements were also utilized to calibrate errors in the detection of the target fixed at the end of the boom. The boom was moved horizontally and vertically in measured movements of 0.5 M to 1 M to calibrate the error between the actual reading of the sensor, the laboratory boom, and the detection model. This methodological framework enabled behavioral analyses, data acquisition, and testing within a controlled environment, thereby supporting the study's robustness. This integration of components allowed precise movement replication, enhancing the investigation's accuracy and scope in a lab setting.

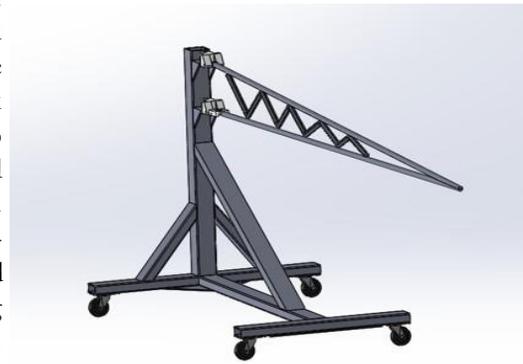

*Figure 4. Experimental Boom designed in lab*

### *LabView Program*

The image-capturing system was integrated with a fanless PC (LEC 2580-711A, Lanner Electronics, Mississauga, ON, Canada) using a LabView program. During the workflow, data collection was seamlessly executed through the LabView program (Figure 5. LabView Program used for data collection).

LabView program utilized fanless pc in addition to dual axis inclinometer sensor (TMS/TMM22, SICK, Minneapolis, USA), place at the starting of the boom to capture angle deviation made by the boom. LabView ties this information as metadata to the image along with the text file. This program allows (a) multiple cameras to be employed, with data recording initiated by a single click of (b) the start button, and data saved in the designated location within (c) the data directory. LabView also offered refinement options for specific camera properties such as (d) buffer time, (e) frame rate, (f) buffer size, (g) reduction factor (image size in pixels), and (h) the size of an individual file. Furthermore, an additional feature includes a (i) Video color format (j) Video type format (m) exposure property of the camera, (n) gain property and (o) distinct section designed for viewing the camera's perspective. This particular zone allows users to position and stabilize the camera precisely as per their specific needs. The LabView program recorded the image and sensor data at 10 Hz.

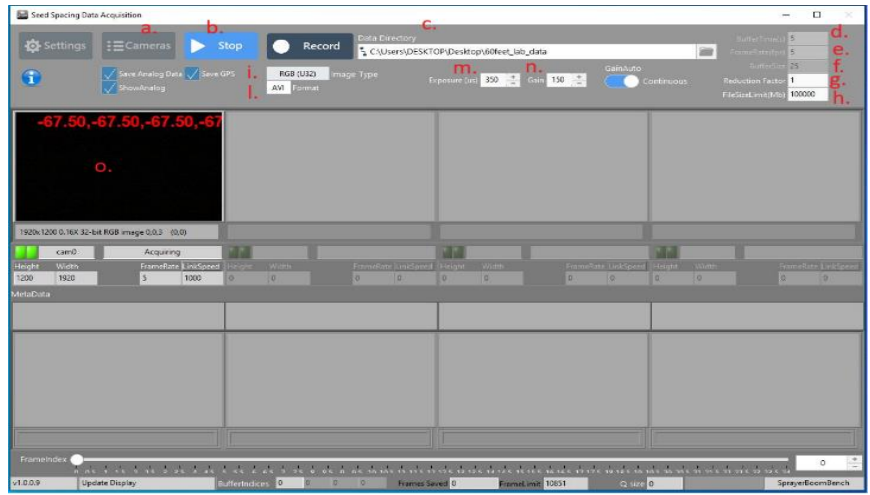

*Figure 5. LabView Program used for data collection*

### Data Collection and Augmentation

The prior stages provided the foundation necessary for deploying the set up to the field, facilitating systematic data collection. In addition to target images collected in the lab using the test stand, an image data set was collected by mounting the selected target in the tip and upper side of the boom (Figure 6. Example). This dataset, comprising 3,490 images, was necessary to include diverse backgrounds (diverse crops and infrastructural elements like buildings and roads, all under differing lighting conditions) for robust detection. Subsequently, the original dataset underwent augmentation processes, incorporating properties such as blur up to 2.5 pixels, exposure adjustments from -25 to 25%, rotations up to 20 degrees, and resizing with auto-adjustment (Figure 7. Data Augmentation examples). This augmentation procedure yielded the final dataset, comprising 9,485 images. The application of augmentation properties on the original dataset, already characterized by considerable diversity, significantly heightened variations in clarity, color, and rotation. The primary objective of this augmentation process was to enhance the model's ability to comprehend the data intricacies, thereby facilitating more effective learning in subsequent phases. Subsequently, the thoroughly augmented dataset underwent annotation using RoboFlow before being fed into the model.

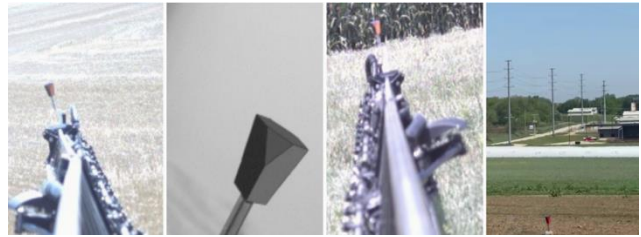

*Figure 6. Example of data collected*

## Training Neural Network

Expediting object detection was achieved through utilization of the DarkNet Model's YOLO models. The decision to utilize the YOLO model was driven by its effectiveness as a single-shot detector. The model segments the image into grids characterized by dimensions N x N, with N being contingent upon factors such as the resolution of the input image and the scale of the anchor boxes. Following this partitioning, a single forward pass through the network is executed for predictions.

The training of YOLO models aimed to achieve optimal results through well-defined neural network configurations. YOLOv7, YOLOv8 and YOLOv11 were chosen due to their recent development and demonstrated advantages in accuracy and inference speed over earlier versions, such as YOLOv5 [22]. Training was conducted under standardized conditions, focusing exclusively on a single class termed "Target," to ensure consistency and comparability. The dataset preparation involved dividing data into training, testing, and validation sets, maintaining a distribution of 74%, 14%, and 12%, respectively. Performance was evaluated based on key metrics, including mean average precision at IoU thresholds of 50% (mAP@50) and 90% (mAP@90), recall, and confidence levels, to provide a robust assessment framework.

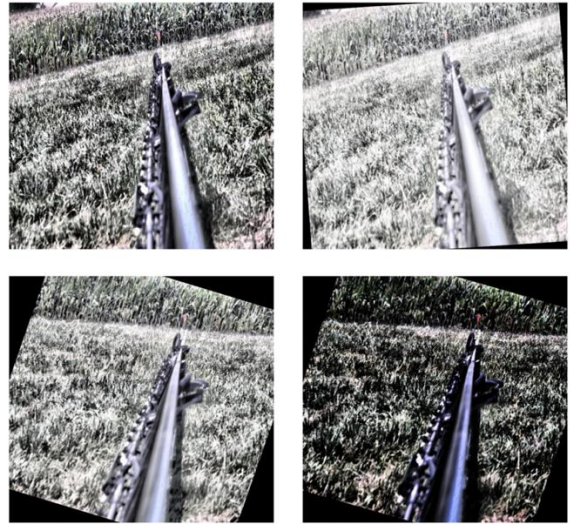

The training process for the data model, conducted with a batch size of 16 and an image resolution of 640 × 640 over 400 epochs, was completed in approximately 45 hours utilizing an NVIDIA Tesla T4 GPU. This duration reflects the computational demands of the specified parameters and hardware configuration.

*Figure 7. Data Augmentation examples*

## RESULTS

This section starts with defining the above-mentioned metrics in the following manner.

*Mean Average Precision (MAP): 50*

This term relates to the calculation of mean average precision, specifically with an Intersection over Union (IoU) of 50% or more between the predicted bounding box and the ground truth box. A noteworthy mAP@.50 score, nearing 1, denotes that the model's predicted bounding boxes display considerable overlap, at least 50%, with the ground truth boxes for a significant number of instances.

*Mean Average Precision (MAP): 90*

This term involves the calculation of mean average precision with an Intersection over Union (IoU) of 90% or more between the predicted bounding box and the ground truth box. A commendable mAP@.90 score, approaching 1, signifies that the model's predicted bounding boxes exhibit significant overlap, at least 90

*Recall*

Recall stands out as a significant metric, quantifying the model's ability to accurately identify relevant instances. It is computed as the ratio of true positives to the total number of actual, relevant instances. The recall score, ranging from 0 to 1, with 1 indicating perfect identification of all relevant instances, is particularly pertinent in object detection scenarios, where a high recall score signifies effective detection of most ground truth objects.

*Confidence Score*

The confidence score, a pivotal metric in object detection, reflects the model's confidence in asserting the presence of an object within a specified bounding box and the precision of the predicted box [21]. When no object is identified in a given cell, the confidence score registers as zero. On the contrary, if an object is present, the confidence score aligns with the Intersection over Union (IoU) between the predicted box and the ground truth. The confidence score is calculated by combining the box confidence (objectness score) and the class confidence. The box confidence represents the model's certainty that a bounding box contains an object, while the class confidence is the probability that the object belongs to a specific class given that an object is present. The final confidence score is the product of these two confidences, reflecting both the model's object detection certainty and its class prediction certainty. During inference, this score does not involve Intersection over Union (IoU) with ground truth, relying solely on the model's predictions [23, 24].

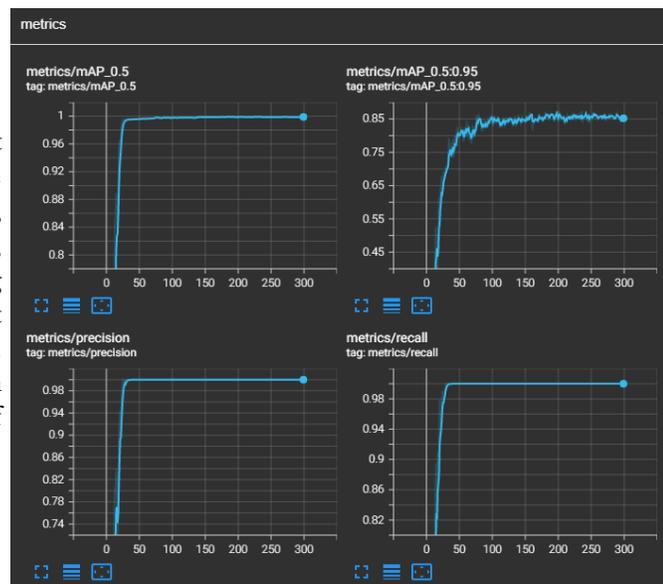

*Figure 8 Results of training for YOLOv7 model*

The Precision score for YOLOv7(Fig. 8) was 0.992%. Recall was also noted as 0.992%. MAP 50 and MAP 90 were 0.982% and 0.847%, respectively. The Precision score for YOLOv8 (Figure 9. Results of training for YOLOv8(n)) was 0.998%. Recall was also noted as 0.998%, and MAP 50 and MAP 90 were 0.995% and 0.797%, respectively. The Precision score for YOLOv11 (Figure 10. Results of training for YOLOv11(n)) was 0.9951%. Recall was also noted as 0.99545%, and MAP 50 and MAP 90 were 0.99495% and 0.77136, respectively.

Higher mAP:50 would be preferred for object detection of a single class in near real-time applications, marking the choice of a neural network as YOLOv8(n).

A thorough evaluation was conducted, involving a comprehensive comparison across multiple dimensions. This assessment (Table 1 Comparison on Confidence on YOLOV7, YOLOV8 and Yolov11) included the Reduction factor, representing image resolution—a reduction factor of 4 indicates a fourfold decrease in image resolution. Various factors, such as image format (color or monochrome) and diverse backgrounds featuring different crops, were systematically examined to assess their impact on confidence levels and inference time, measured in milliseconds.

Simultaneously, a parallel assessment considered diverse YOLO v8 and YOLOv11 versions, spanning nano, small, medium, and large models. Notably, a discernible trend emerged, highlighting a positive correlation between increasing model size and elevated confidence levels, increasing the time taken to infer an object. This observation underscores the significance of model size in influencing both confidence levels and inference speed in the context of the evaluation.

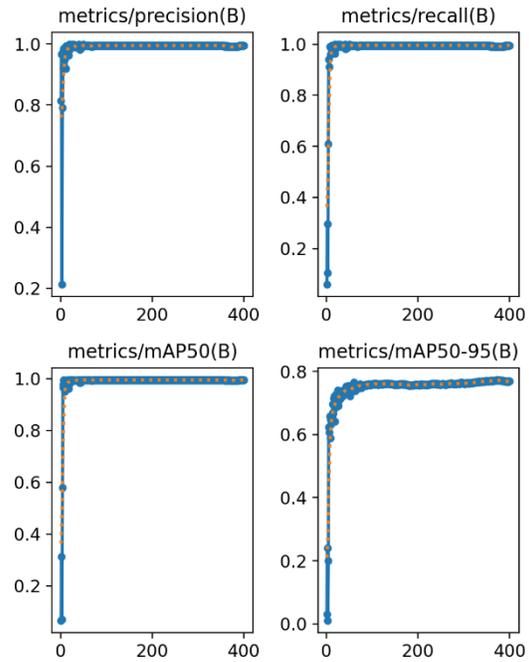

*Figure 9. Results of training for YOLOv8(n) model*

The selection process, guided by the design requirement of higher accuracy, strategically favored certain parameters. Specifically, the YOLOv8(n) model was chosen, coupled with the utilization of the image's color format and a resolution reduction factor of 1 (original size). This deliberate configuration aimed at ensuring a reliable and consistent achievement of optimal accuracy throughout the detection process.

**Fine-tuning**

Subsequent to settling on the neural network based on the specified criteria, an assessment was carried out through a detection test involving around 1,000 images marked by diverse lighting scenarios and varying clarity levels. The run-test implementation involved a thoughtful approach to image inclusion, ensuring a diverse set that prominently featured backgrounds of crops such as Wheat and Corn. The deliberate selection of these backgrounds was guided by a rationale grounded in the recognition that the color similarity prevalent in most crop backgrounds posed a significant consideration (Figure 11 Comparing accuracy of models on wheat and corn crops background). This thorough evaluation brought to light challenges encountered by the network, particularly in effectively identifying the target amid instances of image blurring caused by dirt

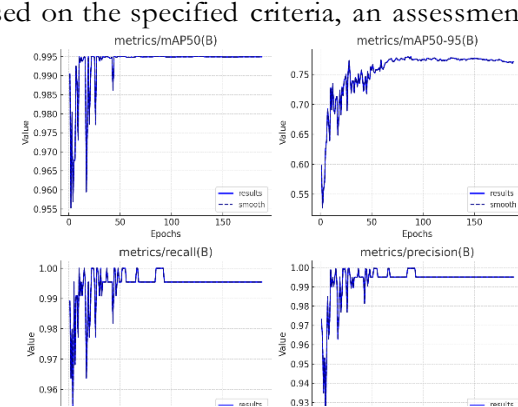

*Figure 10. Results of training for YOLOv11(n) model*

accumulation on the camera lens, notably prevalent during the motion of the sprayer.

*Table 1 Comparison on Confidence on YOLOV7, YOLOV8 and Yolov11*

| Image Resolution | Image Background | Model Parameter | Yolo7 | Yolo8n | Yolo8s | Yolo8m | Yolo8l | Yolo11n | Yolo11s | Yolo11m | Yolo11l |
|---|---|---|---|---|---|---|---|---|---|---|---|
| Reduction 1 (1920x120) | Wheat | Speed (ms) | 1596.7 | 4.1 | 12.3 | 5.8 | 39.5 | 6.0 | 11.0 | 11.1 | 14.8 |
| | | MAP (%) | 78% | 84% | 83% | 79% | 83% | 83% | 78% | 81% | 65% |
| | Corn | Speed (ms) | 1159.3 | 5.0 | 16.0 | 35.4 | 6.0 | 5.4 | 4.8 | 36.8 | 10.0 |
| | | MAP (%) | 71% | 83% | 82% | 77% | 82% | 81% | 75% | 78% | 77% |
| | Color | Speed (ms) | 159.6 | 4.1 | 12.3 | 6.0 | 39.5 | 6.0 | 11.0 | 11.1 | 10.0 |
| | | MAP (%) | 78% | 84% | 83% | 76% | 83% | 83% | 78% | 81% | 77% |
| | Monochrome | Speed (ms) | 1564.5 | 5.0 | 11 | 4.7 | 32.6 | ND | ND | ND | ND |
| | | MAP(%) | 71% (wrong detection) | 83% | 11% | 19% | 56% | ND | ND | ND | ND |
| Reductio 4 (450x300) | Wheat | Speed (ms) | 1681.2 | 7.7 | 8.5 | 8.0 | 10.8 | 9.2 | 11.0 | 6.8 | 30.1 |
| | | MAP(%) | 76% | 82% | 79% | 81% | 81% | 84% | 78% | 78% | 56% |
| | Corn | Speed (ms) | 2638.4 | 4. | 3.4 | 9. | 8. | 7. | 4.8 | 3.5 | 38. |

|  |  |  |  |  |  |  |  |  |  |
|---|---|---|---|---|---|---|---|---|---|
|  |  |  | 0 | .0 | 1 | 7 | 0 | .7 | 7 |
|  | MAP(%) | 67% | 83% | 81% | 83% | 82% | 84% | 75% | 77% | 71% |
| Color | Speed (ms) | 1681.2 | 4.0 | 9.1% | 9.1 | 8.7% | 9.2% | 11.0% | 6.8% | 38.7% |
|  | MAP(%) | 76% | 83% | 83% | 83% | 82% | 84% | 78% | 78% | 71% |
| Monochrome | Speed (ms) | 1496.4 | 5.2 | 9.1 | 9.1 | 11.1 | ND | ND | ND | ND |
|  | MAP(%) | 34% (wrong detection) | 46% | 32% | 32% | 82% | ND | ND | ND | ND |

To effectively address the encountered challenges and amplify the network's efficacy. Images presenting detection challenges, primarily influenced by factors such as blurriness and lens contamination, were systematically compiled. Following this, a detailed annotation process was executed to refine and augment the dataset. This enhanced dataset underwent a subsequent reintroduction into the neural network. The consequential outcomes of this fine-tuning endeavor demonstrated promising results, showcasing a discernible improvement in the network's detection capabilities. The gradual implementation of this refined strategy proved instrumental in substantially strengthening the overall proficiency of the network. As a consequence, there was a notable increase in the detection rate observed across the extensive image dataset.

**Post-Processing**

The successful implementation of detecting blurry images to a good extent allowed us to move further to the post-processing stage, where determining the displacement of the target was conducted with the help of assessing the pixel values. The basic principle to achieve displacement is to compare the pixel coordinate values of the first detected frame to the pixel coordinate values of all the concurrent frames. The process of obtaining actual displacement in distance units, specifically millimeters, necessitates crucial information regarding the area covered by a single pixel. A series of tests were systematically conducted to acquire this information, adhering to the formula: Area under a pixel = Width of the Frame ∕ Number of pixels in the frame. The dimensions of the frame were precisely measured, and the resulting value was divided by the number of pixels to determine the distance covered by an individual pixel. These comprehensive tests spanned varying distances, ranging from 3 to 18 meters, representing the distances from the camera to the target on the sprayer. The calculated value for millimeters covered by one pixel at a distance was established at 0.003196 m at a depth of 18.2 m, serving as a crucial multiplier for pixel values to derive accurate real-life measurements.

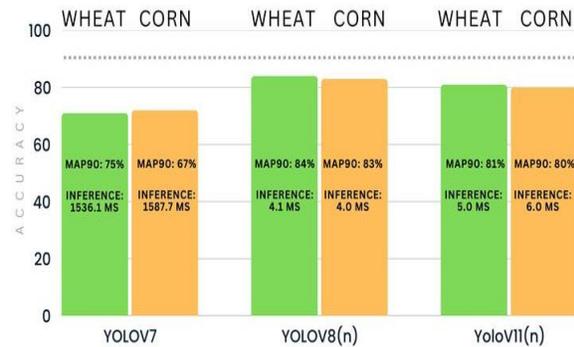

*Figure 11 Comparing accuracy of models on wheat and corn crops*

A further process was followed to store the coordinates of the first detected frame and store it as a reference point (0,0) for the following. Frame's coordinates. Then, these coordinates are subtracted by the first frame coordinate to give the actual displacements in the number of coordinates or pixels. To convert this displacement into real-life measurement terms, we multiply this by multiplier or area covered under a pixel (i.e., is 3.196) ( Figure 12 Flowchart explaining post-Processing steps). This gave us the results for displacement of the boom in millimeters

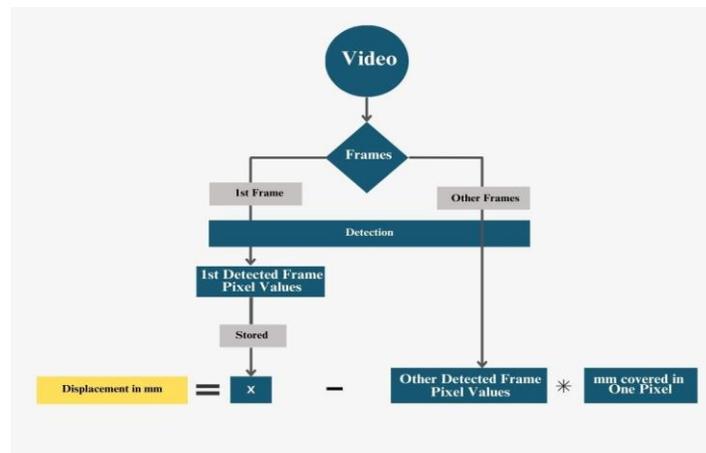

*Figure 12 Flowchart explaining post-Processing steps*

This section primarily concentrates on validating results obtained through a computer vision model by leveraging a hardware sensor, specifically an inclinometer sensor (TMS/TMM 22). Prior to the sensor's application, a series of comprehensive tests were undertaken to discern the sensor's margin of error. The experimental setup involved affixing the sensor to the test boom, maintaining the boom stationary, and recording the sensor's reading fluctuations. This process, repeated 10 times, with each test lasting 10 seconds, aimed to precisely quantify the sensor's error. The observed deflection ranged from a minimum of (-) 0.03 to a maximum of (-) 0.07 degrees. Converting these angular measurements to distance at a range of 18.2 meters yielded minimum and maximum deflections of 0.0095 m and 0.0223mm, respectively. Once the error rate of the sensor was determined, the validation process for comparing the results was done in 2 ways.

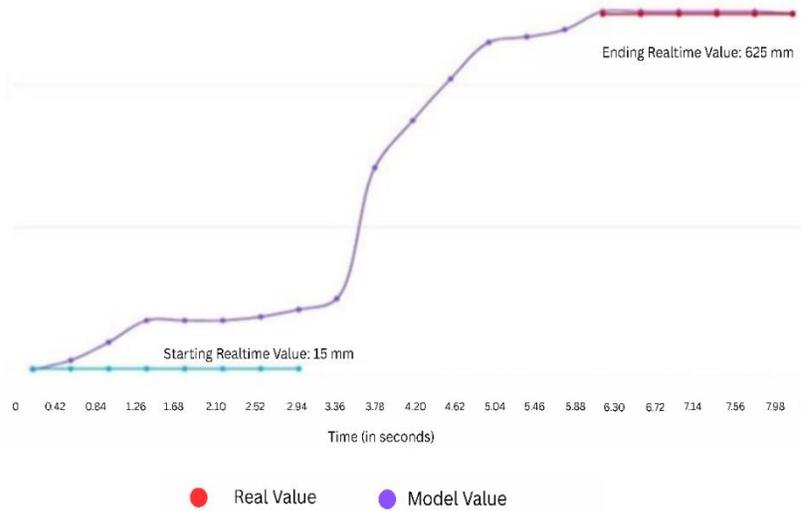

*Figure 13. Manual Validation done against model predicted values recorded at manual reading of 625 mm and model reading of 626 mm*

*Validation done by Experimental Boom*: For this, the initial location of the boom was noted, followed by moving the boom from one end to another and finally marking the end position of the boom. The difference between these two values (initial and final values) gave the distance of movement of the boom, which was then compared to values output by computer vision (Figure 13. Manual Validation done against model predicted values).

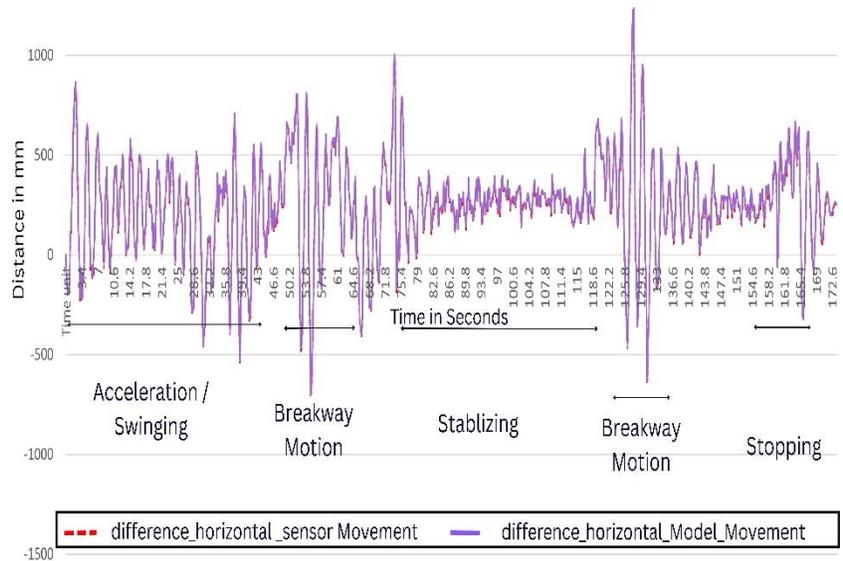

*Figure 14. Sensor Validation done against model predicted values when data recorded at 5 FPS, no value difference of more than 2 cm was recorded*

*Validation done by Sensor*: The foundational strategy adopted herein adhered to the same guiding principle utilized in the computer vision model. Before proceeding, both the LabView image output and sensor output were synchronized at the same FPS to compare the results in a stable environment.

The most recent data from the inclinometer (10 Hz) were compared against the Detection Model (YoloV8(n)). Central to this approach was establishing the first output reading as a reference point for all successive readings.

Given that these readings were denominated in degrees, a methodical conversion process was implemented to translate them into millimeter displacement. The conversion formula, predicated on determining the arc length resulting from the boom's movement, took the form of (Angle / 360) * 2 * 3.14 (pie) * Radius. In this context, the Radius corresponded to the boom's length of 18.2 meters. Simplification of this formula yielded (Angle * 1.046), subsequently multiplied by 304.8 to effect the conversion from feet to millimeters. The outcomes of this conversion were systematically compared against the readings from the computer vision model. (**Figure 14. Sensor Validation done against model predicted values**. In all the tests performed, the error rate was under 0.026 m to the ground truth.

**CONCLUSION**

The study successfully developed an automated computer vision system to quantify the movement of spray booms in agricultural sprayers, addressing a significant contributor to application rate errors. Utilizing YOLO V7, V8, and V11 neural network models, the system could track the boom's movements in real-time field operations with over 90% accuracy. The system's measurements were validated against data from an inclinometer sensor mounted on the boom, showing a close match within 0.026 m. The data gathered by this system can inform design improvements for more stable sprayer booms, leading to greater application accuracy. This represents a significant advancement in addressing the complex challenge of controlling spray booms, particularly given their wide widths, high driving speeds, and the varying terrain.